\documentclass[a4paper]{article}
\usepackage{dx}
\usepackage[T1]{fontenc}
\usepackage{ae,aecompl}
\usepackage{amsfonts}
\usepackage{amsmath}
\usepackage{amsthm}
\usepackage{enumitem}
\usepackage{xspace}

\usepackage{times}

\usepackage{mathnotation}
\usepackage{ensuretext}
\usepackage{booktabs}
\usepackage{multirow}
\usepackage{threeparttable}
\usepackage{graphicx}
\usepackage{colortbl}
\usepackage{caption}

\usepackage{rotating}
\usepackage{pdflscape}

\newcommand{\dpi}{\mathit{dpi}}

\newcommand{\dt}{\mathcal{D}^*}
\newcommand{\md}{\mathcal{D}}
\newcommand{\mD}{\mathbf{D}}

\newcommand{\cost}{\mathsf{cost}}

\newcommand{\ld}{\mathit{ld}}
\newcommand{\Pt}{\mathfrak{P}}
\newcommand{\oracle}{\mathsf{class}}
\newcommand{\ENT}{\mathsf{ENT}}
\newcommand{\SPL}{\mathsf{SPL}}
\newcommand{\KL}{\mathsf{KL}}
\newcommand{\EMCb}{\mathsf{EMCb}}
\newcommand{\MPS}{\mathsf{MPS}}
\newcommand{\BME}{\mathsf{BME}}
\newcommand{\RIO}{\mathsf{RIO}'}
\newcommand{\RND}{\mathsf{RND}}

\newcommand{\tax}{\mathit{ax}}
\newcommand{\mo}{\mathcal{K}}
\newcommand{\mb}{\mathcal{B}}
\newcommand{\Tp}{\mathit{P}}
\newcommand{\Tn}{\mathit{N}}
\newcommand{\tp}{\mathit{p}}
\newcommand{\tn}{\mathit{n}}
\newcommand{\dx}[1]{{\bf D}_{#1}^+}
\newcommand{\dnx}[1]{{\bf D}_{#1}^{-}}
\newcommand{\dz}[1]{{\bf D}_{#1}^0}

\newcommand{\protege}{Prot\'{e}g\'{e}\xspace}

\newtheorem{problem}{Problem}

\newcounter{examplecounter}
\newenvironment{example}{
	\refstepcounter{examplecounter}%
	
	\vspace{7pt}
	\noindent\textbf{Example \arabic{examplecounter}}%
	\quad
}{
	
	\vspace{7pt}
}

\title{Evaluating Active Learning Heuristics for Sequential Diagnosis\thanks{This work was presented at the \emph{Workshop on Principles of Diagnosis 2018 (DX-2018)}, and a version of this work was formally published in the \emph{Proceedings of the International Joint Conference on Rules and Reasoning 2018 (RR+RuleML-2018)} \protect\cite{rodler2018ruleML}.}}
\author%
{%
	Patrick Rodler$^1$ \and 
	Wolfgang Schmid$^1$\\
	$^1$Alpen-Adria Universit\"at Klagenfurt \\
	e-mail: firstname.lastname@aau.at
}

\begin{document}

\maketitle

\begin{abstract}
Given a malfunctioning system, sequential diagnosis aims at identifying the root cause of the failure in terms of abnormally behaving system components. As initial system observations 
usually do not suffice to deterministically pin down just one explanation of the system's misbehavior, additional system measurements can help to differentiate between possible explanations. The goal is to restrict the space of explanations until there is only one (highly probable) explanation left.
To achieve this with a minimal-cost set of measurements, various (active learning) heuristics for selecting the best next measurement have been proposed.

We report preliminary results of extensive ongoing experiments with a set of selection heuristics on real-world diagnosis cases. In particular, we try to answer questions such as ``Is some heuristic always superior to all others?'', ``On which factors does the (relative) performance of the particular heuristics depend?'' or ``Under which circumstances should I use which heuristic?''
%
\end{abstract}

\section{Introduction}
\label{sec:intro}
If the actual behavior of a system does not match its expected behavior, the task of \emph{fault localization} is to determine the \emph{actual diagnosis}, i.e.\ those system components whose faultiness is responsible for the system's improper functioning. A major challenge regarding real-world systems such as hardware, software or knowledge bases in this context is the often enormous number of possible explanations (\emph{diagnoses}) for an observed system failure to begin with. In the presence of the initial observations, each of these diagnoses might correspond to the sought actual one. \emph{Sequential diagnosis (SD)} approaches \cite{dekleer1987} address this fault localization problem by suggesting a sequence of system measurements (e.g.\ electrical measurements in hardware, test cases in software, or questions to a domain expert in knowledge-based systems). The aim of these measurements is to narrow down the space of diagnoses until only one (highly probable) diagnosis persists. Since the realization of such measurements often involves resources in terms of manpower, time or costly equipment, SD methods attempt to minimize the overall accruing measurement costs. Unfortunately, the global optimization of these costs (i.e.\ coming up with the cost-minimal \emph{sequence of measurements} revealing the actual diagnosis) has been proved to be NP-hard \cite{hyafil1976,pattipati1990}. As a result, SD methods have to confine themselves to a local optimization (i.e.\ computing the best \emph{next} measurement). To this end, a one-step lookahead evaluation of measurements (i.e.\ how favorable is the expected situation after making \emph{one} measurement?) proved to be an already very good trade-off between gained information and necessary effort \cite{dekleer1992}. Hence, most, if not all, state-of-the-art SD approaches leverage such one-step-lookahead heuristics.

In this work we focus on one-step lookahead heuristics, called \emph{query selection measures (QSMs)}, that try to optimize measurements formulated as binary true-false tests, called \emph{queries} \cite{Shchekotykhin2012,settles2012}. 
For instance, all of the following can be viewed as queries: a test of a system such as hardware or software \cite{pattipati1990,DBLP:journals/tsmc/ShakeriRPP00,DBLP:journals/jair/FeldmanPG10a,DBLP:conf/aadebug/MateisSWW00}
(given inputs $I$, do we get the expected outputs $O$?), the inspection of system components \cite{heckerman1995decision,DBLP:conf/ijcai/BrodieRMO03} (are all tested components normal?), a question to an expert \cite{Shchekotykhin2012,Rodler2015phd} (is statement $X$ true in domain $D$?), or a probe \cite{dekleer1987} (is the value at measurement point $P$ equal to $v$?).

%
Popular QSMs currently adopted in SD are the \emph{expected information gain} \cite{dekleer1987,DBLP:conf/ijcai/BrodieRMO03,gonzalez2011spectrum,Shchekotykhin2012} and the \emph{split-in-half 
heuristic} \cite{Shchekotykhin2012,Rodler2015phd}. 
A range of new QSMs, most of them 
originally suggested
in the field of active learning \cite{settles2012}, have recently been introduced to SD in our previous work \cite{rodler17dx_activelearning}.
%
%
%
Complementary to the 
analyses in \cite{rodler17dx_activelearning} -- 
mostly addressing the efficient computation of optimal queries wrt.\ QSMs -- we want to bring light to the performance of the QSMs wrt.\ measurement cost throughout an SD session in the present work.
%

For this purpose, we are currently conducting extensive evaluations where we investigate the particular QSMs under varying conditions regarding
\begin{enumerate}[label={(\alph*)},noitemsep]
\item \label{enum:factor_prob-dist} diagnoses probability distributions,
\item quality (meaningfulness) of the probabilities,
\item available diagnostic evidence (size of the diagnoses sample) for query computation, and
\item \label{enum:factor_diag-structure} diagnostic structure (i.e.\ system size; number and cardinality of diagnoses; reasoning complexity)
\end{enumerate}
using real-world diagnosis problems.  
The data of the (already finished) experiments shall be exploited to approach i.a.\ the following questions: 
\begin{itemize}[noitemsep]
\item Do the factors \ref{enum:factor_prob-dist} -- \ref{enum:factor_diag-structure} have an influence on the (relative) performance of the QSMs?
\item Which QSM is preferable under which circumstances?
\item Is there a (clear) winner among the QSMs?
\item What about the difference (variance) between QSM performances under different conditions?
\end{itemize}
The rest of the paper is organized as follows. Sec.~\ref{sec:basics} briefly introduces technical basics wrt.\ SD. Sec.~\ref{sec:QSMs} recaps 
the QSMs used in the experiments. The evaluation setting is described in Sec.~\ref{sec:exp_settings}, and results discussed in Sec.~\ref{sec:exp_results}. 
Sec.~\ref{sec:conclusion} concludes. 

\section{Preliminaries}
\label{sec:basics}
In this section we briefly characterize the basic technical concepts used throughout this work, based on the framework of \cite{Shchekotykhin2012,Rodler2015phd} which is slightly more general \cite{rodler17dx_reducing} than Reiter's theory \cite{Reiter87}. 

\noindent\textbf{Diagnosis Problem Instance (DPI).} A system to be diagnosed, consisting of a set of components $\setof{c_1,\dots,c_k}$, is described by a finite set of logical sentences $\mo \cup \mb$, where $\mo$ (retractable knowledge) characterizes the behavior of the system components, and $\mb$ (correct background knowledge) comprises any additional available 
system knowledge and system observations. More precisely, there is a one-to-one relationship between sentences $\tax_i \in \mo$ and components $c_i$, where $\tax_i$ describes the nominal behavior
of $c_i$.
E.g., if $c_i$ is an OR-gate in a circuit, then $\tax_i := out(c_i) = or(in1(c_i),in2(c_i))$; $\mb$ in this context might subsume sentences stating, e.g., which components are connected by wires, 
or observed outputs of the circuit. The inclusion of a sentence $\tax_i$ in $\mo$ corresponds to the implicit assumption that $c_i$ is healthy. Evidence about the system behavior is captured by sets of positive ($\Tp$) and negative ($\Tn$) measurements \cite{Reiter87,dekleer1987,DBLP:journals/ai/FelfernigFJS04}. Each measurement is a logical sentence; positive ones $\tp\in\Tp$ must be true and negative ones $\tn\in\Tn$ must not be true. 
We call $\tuple{\mo,\mb,\Tp,\Tn}$ a \emph{diagnosis problem instance (DPI)}. 

\noindent\textbf{Diagnoses.} Given that the system description along with the positive measurements (under the 
assumption $\mo$ that all components are healthy) is inconsistent, i.e.\ $\mo \cup \mb \cup \Tp \models \bot$, or some negative measurement is entailed, i.e.\ $\mo \cup \mb \cup \Tp \models \tn$ for some $\tn \in \Tn$, some assumption(s) about the healthiness of components, i.e.\ some sentences in $\mo$, must be retracted. We call such a set of sentences $\md \subseteq \mo$ a \emph{diagnosis} for the DPI $\tuple{\mo,\mb,\Tp,\Tn}$ iff $(\mo \setminus \md) \cup \mb \cup \Tp \not\models x$ for all $x \in \Tn \cup \setof{\bot}$. We say that $\md$ is a \emph{minimal diagnosis} for $\dpi$ iff there is no diagnosis $\md' \subset \md$ for $\dpi$. A sample $\mD$ of minimal diagnoses, the \emph{leading diagnoses}, is often used as an input for measurement selection algorithms, serving as a basis for measurement quality assessment. We call $\dt$ \emph{the actual diagnosis} iff all $\tax \in \dt$ are faulty and all $\tax \in \mo\setminus\dt$ are healthy.

If component fault probabilities are available -- formalized as $p(\tax_i)$ for $\tax_i \in \mo$ and defining how likely is it that the nominal behavior description $\tax_i$ of $c_i$ does not apply -- probabilities of diagnoses $\md \in \mD$ (of being the actual diagnosis) can be computed as $$p(\md) = \prod_{\tax \in \md} p(\tax) \prod_{\tax \in \mo\setminus\md} (1 - p(\tax)) $$
and updated by means of Bayes' Rule (see \cite[p.~130]{Rodler2015phd}) after a new measurement is added. 
Sometimes $p(\tax)$ for $\tax \in \mo$ might not be directly given, e.g., if components have an internal structure in that they constitute an aggregation of several sub-components $sc_i$. In such a case, $p(\tax)$ can be computed from fault probabilities $p(sc_i)$ of sub-components as \cite{Shchekotykhin2012}:
$$ p(\tax) = 1-\prod_{sc_i \text{ occurs in } \tax} (1-p(sc_i))$$  

\noindent\textbf{Queries and Q-Partition.} Let $\mD$ be a set of leading diagnoses for $\dpi = \tuple{\mo,\mb,\Tp,\Tn}$. A \emph{query} (wrt.\ $\mD$) is a logical sentence $q$ that rules out at least one diagnosis in $\mD$, both if $q$ is classified as a positive measurement ($\Tp \gets \Tp \cup \setof{q}$), and if $q$ is classified as negative measurement ($\Tn \gets \Tn \cup \setof{q}$). That is, at least one $\md_i \in \mD$ is not a diagnosis for $\tuple{\mo,\mb,\Tp\cup\setof{q},\Tn}$ and at least one diagnosis $\md_j \in \mD$ is not a diagnosis for $\tuple{\mo,\mb,\Tp,\Tn\cup\setof{q}}$. The classification of a query $q$ to either $\Tp$ or $\Tn$ is accomplished by an oracle, e.g.\ an engineer performing measurements or a domain expert answering questions. The \emph{oracle} is a function $\oracle: \mathbf{Q} \to \setof{\Tp,\Tn}$ where $\mathbf{Q}$ is the relevant query space (a set of logical sentences). 

An expedient tool towards the verification and goodness estimation of query candidates $q$ is the notion of a q-partition. 
Namely, every logical sentence $q$ partitions a set of leading diagnoses $\mD$ into three subsets: 
\begin{itemize}[]
\item $\dx{q}$: includes all $\md \in \mD$ where $\md$ is not a diagnosis for $\tuple{\mo,\mb,\Tp,\Tn\cup\setof{q}}$\\
(diagnoses predicting that $q$ is a positive measurement)
\item $\dnx{q}$: includes all $\md \in \mD$ where $\md$ is not a diagnosis for $\tuple{\mo,\mb,\Tp\cup\setof{q},\Tn}$\\
(diagnoses predicting that $q$ is a negative measurement)
\item $\dz{q} = \mD \setminus (\dx{q} \cup \dnx{q})$: includes all $\md \in \mD$ where $\md$ is a diagnosis for both $\tuple{\mo,\mb,\Tp\cup\setof{q},\Tn}$ and $\tuple{\mo,\mb,\Tp,\Tn\cup\setof{q}}$\\
(\emph{uncommitted diagnoses}, no prediction about $q$)
\end{itemize}
A $3$-partition $\Pt$ of $\mD$ is called \emph{q-partition (QP)} iff there is a query $q$ for $\mD$ such that $\Pt = \tuple{\dx{q},\dnx{q},\dz{q}}$.
According to the definition of a query, it holds that $q$ is a query iff both $\dx{q}$ and $\dnx{q}$ are non-empty sets. This fact can be taken advantage of for \emph{query verification}. 

Coupled with diagnoses probabilities, the QP provides useful hints \cite{rodler17dx_activelearning} about \emph{query quality} in that it enables to
\begin{enumerate}[label={(\arabic*)},noitemsep]
\item \label{enum:qqp:strong_query} test whether $q$ is a \emph{strong query}, i.e.\ one without uncommitted diagnoses ($\dz{q}=\emptyset$),
\item estimate the impact $q$'s classification $\oracle(q)$ has in terms of diagnoses elimination (potential a-posteriori change of the diagnoses space), and
\item \label{enum:qqp:prob_of_pos_neg_answer} assess the probability of $q$'s positive and negative classification 
(e.g.\ to compute the uncertainty of $q$).
\end{enumerate}
As per \cite{dekleer1987}, given a set of leading diagnoses $\mD$, we estimate 
$p(\oracle(q) = \Tp) = p(\dx{q})+\frac{1}{2}p(\dz{q})$
and 
$p(\oracle(q) = \Tn) = p(\dnx{q})+\frac{1}{2}p(\dz{q})$
where $p(\mD_q^{X})=\sum_{\md\in\mD_q^{X}}p(\md)$
for $X \in \setof{+,-,0}$ and $p(\md)$ for $\md \in \mD$ is the probability of $\md$ normalized over $\mD$ (i.e.\ $\sum_{\md\in\mD} p(\md) = 1$).

\noindent\textbf{Sequential Diagnosis.} Formally, the (optimal) SD problem can be stated as follows:
\begin{problem}[(Optimal) SD]\label{prob:opt_SD}
\textbf{Given:} A DPI $\tuple{\mo,\mb,\Tp,\Tn}$.\\
\textbf{Find:} An (optimal-cost) set of measurements $\Tp' \cup \Tn'$ such that there is only a single minimal diagnosis for $\tuple{\mo,\mb,\Tp\cup\Tp',\Tn\cup\Tn'}$.
\end{problem}

\noindent\textbf{Query Selection Measures (QSMs).} The said query properties \ref{enum:qqp:strong_query} -- \ref{enum:qqp:prob_of_pos_neg_answer} characterized by the QP are essentially what QSMs take into account to quantitatively rate the query quality. Formally, a QSM is a function $m: \mathbf{Q} \to \mathbb{R}$ that assigns a value $m(q)$ to each query $q \in \mathbf{Q}$. All QSMs are heuristics towards Optimal SD (Problem~\ref{prob:opt_SD}). That is, their goal is to minimize the expected cost $\sum_{\md} p(\md) \cost(\md)$ of locating the actual diagnosis $\dt$. At this, $\cost(\md)$ is usually
conceived of as the sum of individual query costs over all queries required to unambiguously isolate $\md$. For the purpose of this paper we assume $\cost(\md)$ represents the number of queries to isolate $\dt$ (all queries assumed equally costly). 

%

\renewcommand{\arraystretch}{1.5}
\setlength{\tabcolsep}{10pt}
\begin{table}[t] 
\scriptsize
\begin{center}
\begin{tabular}{@{}lcc@{}}
\toprule
QSM $m$  			& $m(q)$ & opt. \\
\midrule
$\ENT$ 		& $\sum_{c\in\setof{\Tp,\Tn}} p(\oracle(q)=c) \log_2 p(\oracle(q)=c)$     & $\searrow$ 	\\
$\SPL$ 		& $\left|\, |\dx{q}| - |\dnx{q}| \,\right|$  & $\searrow$ \\
$\mathsf{KL}$ 		& $-\sum_{X\in\setof{\dx{q},\dnx{q}}}\frac{|X|}{|\dx{q} \cup \dnx{q}|} \log_2 \frac{p(X)}{p(\dx{q} \cup \dnx{q})}$ & $\nearrow$ \\
$\EMCb$		& $p(\oracle(q)=\Tp) |\dnx{q}| + p(\oracle(q)=\Tn) |\dx{q}|$ & $\nearrow$ 	\\
$\MPS$ 		& $p(\mD_{q,\min})$ if $|\mD_{q,\min}|=1$,$\quad 0$ else  \;$_{\textbf{1)}}$  & $\nearrow$ \\
$\BME$ 		& $|\mD_{q,p,\min}|$ \;$_{\textbf{2)}}$ & $\nearrow$ 	\\
$\RIO$ 		& $\frac{\mathsf{ENT}(Q)}{2}+\mD_{q,n}$ \;$_{\textbf{3)}}$ & $\searrow$ \\
\bottomrule
\end{tabular}
\end{center}
\renewcommand{\arraystretch}{1.4}

\vspace{-5pt}
\centering
\begin{tabular}{@{}lp{6.7cm}@{}}
\textbf{1):} & $\mD_{q,\min} := \argmin_{X\in\setof{\dx{q},\dnx{q}}}(|X|)$. \\

\textbf{2):} & $\mD_{q,p,\min} := \begin{cases} \dnx{q}  &\mbox{if } p(\dnx{q}) < p(\dx{q}) \\ 
\dx{q} & \mbox{if } p(\dx{q}) < p(\dnx{q}) \\
0 & \mbox{else}
\end{cases}$ \\
\textbf{3):} & $\mD_{q,n}:= \begin{cases} {c}_q - n  &\mbox{if } {c}_q \geq n \\ 
|\mD| & \mbox{else } \end{cases}$ \\ 
& where ${c}_q := \min\{|\dx{q}|,|\dnx{q}|\}$	and $n$ denotes the minimal number of diagnoses the selected query must eliminate 		\cite{Rodler2013}\\	
\end{tabular}
\vspace{-0pt}
\caption{\small 
	(\protect\cite[Tab.~2]{rodler17dx_activelearning}) 
	QSM designators (col.~1) and according functions $m(q)$ (col.~2). Col.~3 indicates whether QSM is optimized by maximization ($\nearrow$) or minimization ($\searrow$).
 }
\label{tab:QSMs}
\end{table}

\section{The Evaluated Heuristics}
\label{sec:QSMs}
In this section we briefly 
revisit and explain the QSMs -- originally introduced in other works -- we use in our experiments. These include the 
``classical'' ones \cite{dekleer1987,Shchekotykhin2012} frequently
used in SD (cf.\ Sec.~\ref{sec:intro}) and the newer ones proposed in \cite{Rodler2013} and 
\cite{rodler17dx_activelearning} and discussed in-depth in 
\cite{DBLP:journals/corr/Rodler16a}. Since we employ a query computation and selection method \cite{rodler17dx_queries} that guarantees 
to produce only (the more favorable, cf.\ \cite[Sec.~2.4.1]{rodler_jair-submission_2017}) strong queries, \cite[Tab.~3]{rodler17dx_activelearning} tells us that we have to deal with seven (non-equivalent) QSMs in this case. We next illustrate the rough idea behind these heuristics, listed in Tab.~\ref{tab:QSMs}.

	\noindent\textbf{Information Gain $\ENT$:} Chooses a query with the highest expected information gain or, equivalently, with the lowest expected posterior entropy wrt.\ the diagnoses set $\mD$. 
	As derived in \cite{dekleer1987}, $\ENT(q)$ is the better, the closer the probabilities for positive and negative classification of $q$ are to $0.5$ (cf.\ formula in 
	Tab.~\ref{tab:QSMs}). 
	
	\noindent\textbf{Split-In-Half $\SPL$:} Chooses a query $q$ whose QP best splits the diagnoses set $\mD$ in half, i.e.\ where both $|\dx{q}|$ and $|\dnx{q}|$ are closest to $\frac{1}{2}|\mD|$. Intuitively, an optimal $q$ wrt.\ $\SPL$ guarantees that a half of the (known) diagnoses are eliminated by querying $q$'s classification.
	
	\noindent\textbf{Kullback-Leibler Divergence $\KL$:}  Chooses a query 
	with
	largest average disagreement between query-classification predictions of single diagnoses $\md \in \mD$ and the consensus (prediction) of all $\md\in\mD$, based on an information-theoretic measure of the difference between two probability distributions \cite{settles2012}. As demonstrated in \cite[Prop.~26]{DBLP:journals/corr/Rodler16a}, this QSM can be represented in terms of the formula given in Tab.~\ref{tab:QSMs}. 
	
	\noindent\textbf{Expected Model Change $\EMCb$:} Chooses a query for which the expected number of invalidated diagnoses in $\mD$ is maximized. 
	
	\noindent\textbf{Most Probable Singleton $\MPS$:} 
	Chooses a query $q$ for which the minimum-cardinality set among $\setof{\dx{q},\dnx{q}}$ is a singleton $\setof{\md}$ where $\md$ 
	has maximal probability.
	Intuitively, $\MPS$ seeks to eliminate, with a maximal probability, the maximal possible number of $|\mD|-1$ diagnoses in $\mD$. 
	
	\noindent\textbf{Biased Maximal Elimination $\BME$:}
	Chooses a query with a bias (probability > 0.5) towards one classification ($\Tp$ or $\Tn$) such that this more likely classification rules out an as high as possible number of diagnoses in $\mD$. 
	
	\noindent\textbf{Risk Optimization $\RIO$:} Chooses a query with optimal information gain ($\ENT$-value) among those that, in the worst case, eliminate (at least) $n \leq \frac{1}{2}|\mD|$ diagnoses in $\mD$. At this, the parameter $n$ is learned by reinforcement based on the diagnoses elimination performance achieved so far during an SD session.\footnote{Note, we consider the slightly modified version $\RIO$ of the original QSM $\mathsf{RIO}$ \cite{Rodler2013}, as suggested in \cite{rodler17dx_activelearning}.}

In addition to these informed QSMs, we used a random QSM in our evaluations as a baseline:

	\noindent\textbf{Random $\RND$:} Samples one element uniformly at random from the considered query space $\mathbf{Q}$.

\renewcommand{\arraystretch}{1.1}
\setlength{\tabcolsep}{5pt}
\begin{table}[t] 
\scriptsize
\begin{center}
\begin{tabular}{@{}lllrr@{}}
\toprule
$q$ & $\dx{q}$ & $\dnx{q}$ & $p(\dx{q})$ & $p(\dnx{q})$ \\
\midrule
$\tax_1$ & $\setof{\md_1,\md_2,\md_3,\md_4,\md_6}$ & $\setof{\md_5}$ & 0.59 & 0.41 \\
$\tax_2$ & $\setof{\md_5,\md_6}$ & $\setof{\md_1,\md_2,\md_3,\md_4}$ & 0.45 & 0.55 \\
$\tax_3$ & $\setof{\md_2,\md_3,\md_4,\md_5}$ & $\setof{\md_1,\md_6}$ & 0.95 & 0.05 \\
$\tax_4$ & $\setof{\md_1,\md_2,\md_3,\md_4}$ & $\setof{\md_5,\md_6}$ & 0.55 & 0.45 \\
$\tax_5$ & $\setof{\md_1,\md_3,\md_4,\md_5,\md_6}$ & $\setof{\md_2}$ & 0.67 & 0.33 \\
$\tax_6$ & $\setof{\md_1,\md_2,\md_4,\md_5,\md_6}$ & $\setof{\md_3}$ & 0.86 & 0.14 \\
$\tax_7$ & $\setof{\md_1,\md_2,\md_3}$  & $\setof{\md_4,\md_5,\md_6}$ & 0.48 & 0.52 \\
\bottomrule
\end{tabular}
\end{center}
\caption{\small Q-partitions for the queries $q_i := \tax_i$ in Ex.~\ref{ex:QSMs}.}
\label{tab:example_QPs}
\end{table}

We next illustrate these different selection principles:
\begin{example}\label{ex:QSMs}
Consider a DPI (cf.\ \cite[Tab.~1+2]{DBLP:journals/corr/Rodler16a}) with $\mo = \setof{1,\dots,7}$ (where numbers $i$ denote sentences $\tax_i$) which gives rise to the minimal diagnoses $\mD$ given by $$\setof{\md_1,\dots,\md_6} = \setof{[2,3],[2,5],[2,6],[2,7],[1,4,7],[3,4,7]}$$  and diagnoses probabilities
$$\setof{p(\md_1),\dots,p(\md_6)} = \setof{0.01,0.33,0.14,0.07,0.41,0.04}$$
Let, for simplicity, the possible queries be direct tests of components $c_i$. That is, there are seven queries $q_i := \tax_i$ ($i=1,\dots,7$) where their QPs 
are shown in Tab.~\ref{tab:example_QPs}. Then, the query choice of the discussed QSMs is as follows:
\begin{itemize}[noitemsep]
\item $\ENT$: $q_7$, as $p(\dx{q_7})$ and $p(\dnx{q_7})$ are closest to $0.5$.
\item $\SPL$: $q_7$, as $|\dx{q_7}|$ and $|\dnx{q_7}|$ are equal to $\frac{|\mD|}{2}=3$. 
\item $\KL$: $q_3$, as $\KL(q_3)=1.48$ is maximal over all queries.
\item $\EMCb$: $q_7$, as the expected number of eliminated diagnoses is $3$ (and lower for all other queries).
\item $\MPS$: $q_1$, as  $|\dnx{q_1}|=|\setof{\md_5}|=1$ and $p(\md_5)=0.41 > 0.33 (\text{cf. }  q_5) > 0.14 (\text{cf. } q_6)$.
\item $\BME$: $q_7$, as it has a $\BME$-value of $3$ (larger than the value of all other queries).
\item $\RIO$ (with $n=2$): $q_2$ or $q_4$, as these are the queries with best $\ENT$-value among all queries ($q2,q3,q4$) which eliminate $n$ diagnoses in $\mD$ in the worst case.
\end{itemize}
\end{example}
\setlength{\tabcolsep}{8pt}
\begin{table}[t]
\begin{threeparttable}[t]
\renewcommand\arraystretch{1.2}
\scriptsize
\centering
\begin{tabular}{@{}llrll@{\kern3pt}} 
\toprule
$j$ &KB $\mo_j$				& $|\mo_j|$& reasoning complexity \tnote{a} 		& \#D/min/max \tnote{b} \\ \midrule
1 & University (U) \tnote{c}   		 & 50 		& $\mathcal{SOIN}^{(D)}$& 90/3/4      \\		
2 & MiniTambis (M) \tnote{c}			& 173 		& $\mathcal{ALCN}$ 		& 48/3/3	  \\
3 & Transportation (T) \tnote{c}		& 1300 		& $\mathcal{ALCH}^{(D)}$& 1782/6/9	  \\
4 & Economy (E) \tnote{c}			& 1781 		& $\mathcal{ALCH}^{(D)}$& 864/4/8     \\
\bottomrule
\end{tabular}
\begin{tablenotes}
\item[a] Description Logic expressivity, cf.\ \cite[p.~525ff.]{DLHandbook}.
\item[b] \#D, min, max denote the number, the min.\ and max.\ size of diagnoses \emph{for the initial DPI} (the actual diagnoses search space explored in an SD session might be substantially larger due to a change of the DPI after each measurement addition; also, diagnoses sizes can only increase during an SD session \cite{Reiter87}). 
\item[c] Sufficiently complex systems (\#D $\geq 40$) used in \cite{Shchekotykhin2012}.
\end{tablenotes}
\vspace{-7pt}
\caption{\small Dataset used in experiments.}
\label{tab:dataset}
\end{threeparttable}
\end{table}

\section{Experimental Settings}
\label{sec:exp_settings}
\textbf{The Dataset.} 
%
Tab.~\ref{tab:dataset} depicts the (part of the overall) dataset investigated in the already finished experiments.
The underlying systems are faulty (inconsistent) real-world knowledge bases (KBs). The DPIs $\dpi_j$ we extracted from these KBs $\mo_j$ were $\tuple{\mo_j,\emptyset,\emptyset,\emptyset}$, 
i.e.\ the background $\mb$, positive ($\Tp$) and negative ($\Tn$) measurements were (initially) empty.
%

Please note that using KBs as test cases does not restrict the generality \cite{rodler17dx_reducing} of the results -- any MBD problem can be modeled as a faulty KB. It is rather the diagnostic structure (e.g.\ system size, number and size 
of diagnoses, reasoning complexity) and the used meta information (e.g.\ probabilities) that have a major influence. The diagnostic structure for the used test cases is given in 
Tab.~\ref{tab:dataset}. 

Concerning the 
system components, 
note that each $\tax \in \mo$ is (interpreted as) describing a complex component, consisting of sub-components, where the latter are given by the logical operators occurring in $\tax$, as done in \cite{Shchekotykhin2012}. This interpretation helped us to obtain a self-consistent ascription of component and diagnoses probabilities by specifying the probabilities of the sub-components (cf.\ Sec.~\ref{sec:basics}).

\noindent\textbf{The Factors.} To test the behavior and robustness of the discussed QSMs under various scenarios, we -- in addition to the DPI -- 
varied the following factors in our experiments:
\begin{enumerate}[label={(F\arabic*)},noitemsep]
\item the type of (sub-component) probability distribution 
(\emph{non-biased}, \emph{moderately biased}, \emph{strongly biased}); 
\item $3$ different random choices of assigned probabilities for each distribution type (to average out potential peculiarities of a specific probability assignment),
\item the plausibility of the probabilities (simulated by \emph{plausible}, \emph{random}, \emph{implausible} oracle behavior),
\item the amount of information available for query selection (number of leading diagnoses $\ld \in \setof{6,10,14}$), and
\item the actual diagnosis $\dt$ (i.e.\ the target solution of the SD sessions). 
\end{enumerate}
\emph{Ad (F1)}: Let $\mathit{SC}$ denote the set of all sub-components occurring in some $\tax\in\mo$ and $E_{\lambda}(x) = \lambda e^{-\lambda x}$ the probability density function of the exponential distribution. Three probability distribution types were modeled, by assigning to each sub-component in $\mathit{SC}\dots$
\begin{itemize}[noitemsep]
\item \emph{all-equal (EQ)}: $\dots$an equal (random) value $r \in [0,1]$
\item \emph{moderately biased (MOD)}: $\dots$the probability $E_{\lambda}(x_i)$ for a random $x_i \in [i-\frac{1}{2},i+\frac{1}{2})$ where $i$ is randomly chosen (without replacement) from $\setof{1,\dots,|\mathit{SC}|}$ and $\lambda := 0.5$ (cf.\ \cite{Shchekotykhin2012})
\item \emph{strongly biased (STR)}: $\dots$the probability $E_{\lambda}(x_i)$ for a random $x_i \in [i-\frac{1}{2},i+\frac{1}{2})$ where $i$ is randomly chosen (without replacement) from $\setof{1,\dots,|\mathit{SC}|}$ and $\lambda := 1.75$ (cf.\ \cite{Shchekotykhin2012})
\end{itemize}
Intuitively, one can view both MOD and STR to (1)~precompute a sequence $p_1 > \dots > p_{|\mathit{SC}|}$ of values in $(0,1)$ where, on average, 
the ratio between each value $p_i$ and the next smaller one $p_{i+1}$ is $p_i/p_{i+1} = e^\lambda$, i.e.\ $\approx 1.6$ for MOD and $\approx 5.8$ for STR, and (2)~assign to each $sc\in\mathit{SC}$ a randomly chosen probability $p_i$ from this sequence without replacement. 
Hence, if sorted from large to small, the sub-component probabilities are completely uniform for EQ (\emph{no bias}), moderately descending for MOD (\emph{moderate bias}) and steeply descending for STR (\emph{strong bias}). 

For instance, EQ could model a situation where a novel device gets defect or a novice knowledge/software engineer obtains faulty code, and there is no relevant fault information about device parts or code at hand. On the other hand, MOD can be interpreted to simulate a moderate tendency 
in the fault information,
i.e.\ a non-negligible number of (sub-)components that have a non-negligible fault probability. 
For example, assume a car mechanic who knows from experience that, say, a dozen of parts are mostly responsible for failures whereas all others must be very rarely replaced. 
%
%
STR reflects cases where the differences in fault likeliness are substantial, i.e.\ very few (sub-)components have a non-negligible probability of being abnormal, whereas most of the (sub-)components are practically always nominal. An example is a knowledge engineer that, in the past, has made almost only errors regarding quantifiers. 

\noindent\emph{Ad (F3)}: Let $q$ be a query with $p(\oracle(q)=\Tp) = x$ (cf.\ Sec.~\ref{sec:basics}). The plausibility of the given probabilistic information was modeled by different oracle functions $\oracle$, simulating different strategies of query classification:  
\begin{itemize}[noitemsep]
\item \emph{plausible}: classify $q$ to $\Tp$ with probability $x$ 
\item \emph{random}: classify $q$ to $\Tp$ with probability $0.5$ 
\item \emph{implausible}: classify $q$ to $\Tp$ with probability $1-x$ 
\end{itemize}
Recall (Sec.~\ref{sec:basics}), $p(\oracle(q)=\Tn) = 1-p(\oracle(q)=\Tp)$. 
The idea is that, given (un)reasonable fault information, the estimated query classification probabilities should be good (bad) approximations of the real likeliness of getting a respective outcome. The plausible scenario reflects the case where given probabilities are useful and provide a rational bias, 
e.g.\ for mass products that have been on the market for a long time. The random strategy
aims at estimating the average number of queries needed to pin down $\dt$, 
assuming we cannot make useful predictions about the oracle. The implausible strategy represents a 
misleading fault model, where probabilities turn out to be opposite to what the given information suggests, e.g., when using subjective estimates or historical data that does not apply to the present scenario.
As QSMs utilize fault information for query suggestion, we want to assess their robustness under changing fault information quality \cite{Shchekotykhin2012,Rodler2013}. 

\noindent\emph{Ad (F5)}: We specified the target solution $\dt$ in the different SD sessions implicitly through the oracle answer strategies, see (F3). That is, each SD session continued until positive ($\Tp'$) and negative ($\Tn'$) measurements were collected such that there was just a single minimal diagnosis for the DPI $\tuple{\mo_j,\emptyset,\Tp',\Tn'}$ (resulting from the initial DPI by adding $\Tp'$ and $\Tn'$, cf.\ Problem~\ref{prob:opt_SD}).
This implicit definition of $\dt$ has the advantage of higher generality and closeness to reality over a prior explicit fixation of some $\dt$ among the minimal diagnoses for the initial DPI $\tuple{\mo_j,\emptyset,\emptyset,\emptyset}$ \cite{Shchekotykhin2012}. Because, in the latter case
only one specific
class of SD problems is considered, namely those where the actual solution is already a minimal diagnosis for the start problem. In practice, it might well be the case that the actual solution is a superset of an initial minimal diagnosis.
%

\noindent\textbf{The Tests.} For each of the DPIs $\dpi_1,\dots,\dpi_4$, for each of the $8$ QSMs explicated in Sec.~\ref{sec:QSMs}, and for each of the $3^4$ factor level combinations of factors (F1) -- (F4) we performed $20$ SD sessions. Factor (F5) was implicitly varied in these $20$ runs through the randomized oracle behavior (F3), yielding in most cases a different $\dt$. In case some $\dt$ happened to occur repeatedly throughout the $20$ sessions, we discarded such duplicate runs.\footnote{For logs, etc.\ see http://isbi.aau.at/ontodebug/evaluation.}

\section{Experimental Results}
\label{sec:exp_results}
\subsection{Representation}
The obtained results
are shown by Figures~\ref{fig:eval_university_overall} -- \ref{fig:eval_economy} which graph the number of queries required by the tested QSMs until $\dt$ could be isolated. At this, the green / yellow / red bars depict the situation of a plausibly / randomly / implausibly answering oracle (F3). Each bar represents an average over (up to) $20$ sequential diagnosis sessions (F5) and $3$ random choices of probabilities (F2). Each figure summarizes the results for one case in Tab.~\ref{tab:dataset}; the plots for the U and T cases are more comprehensive, including all combinations of factor levels for (F1), (F2) and (F4), whereas the depictions of M and E are kept shorter due to space restrictions, showing only the $\ld = 10$ case of (F4) for all settings of (F1) and (F2). Along the x-axes of the figures we have the $8$ different QSMs, grouped by manifestations of factor 
(F4)
in Figures~\ref{fig:eval_university_overall} and \ref{fig:eval_transportation_overall}, and by instantiations of factor 
(F1)
in Figures~\ref{fig:eval_minitambis} and \ref{fig:eval_economy}. 

\subsection{Observations}
Interesting gained insights are discussed next. 

\noindent\textbf{Is there a Clear Winner?}
This question can be answered negatively pretty clearly.
For instance, have a look at the MOD, $\ld = 14$ case in Fig.~\ref{fig:eval_university_overall}. Here we see that $\MPS$ performs really good compared to all other QSMs for all oracle types. In fact, it is better than all others in the plausible and random configurations, and loses just narrowly against $\RND$ given implausible answers. However, if we draw our attention to, e.g., the EQ case in the same figure, we recognize that $\MPS$ comes off significantly worse than other heuristics under a plausible oracle behavior. Similar argumentations apply for all other potential winner QSMs. For $\ld = 10$,  Tab.~\ref{tab:eval_best_QSM_per_setting}, which lists the best QSMs in all the different settings we investigated, confirms that there is no single best QSM. 

\noindent\textbf{Sensitivity to Fault Information.}
That there is no QSM which always outmatches all others is not a great surprise, as we evaluate under various types of given probabilistic information $p(.)$ and the different measures exploit $p(.)$ to a different extent when selecting a query. As a result, we can observe probability-independent QSMs such as $\SPL$ outperform (lose against) strongly probability-reliant ones such as $\ENT$ in situations where the fault information is wrongly (reasonably) biased, e.g., see the implausible (plausible) cases for MOD and STR in Figures~\ref{fig:eval_university_overall} and \ref{fig:eval_transportation_overall}.
So, e.g., $\SPL$ can never benefit from high-quality meta information about faults, but cannot effect a significant overhead given low-quality probabilities either. The behavior of, e.g., $\ENT$, is diametrically opposite. To verify this, check the difference between the green and red bars for both $\SPL$ and $\ENT$ for MOD and STR; for $\SPL$ they are hardly different at all, whereas for $\ENT$ they diverge rapidly as we raise the bias (EQ $\to$ MOD $\to$ STR) in the underlying distribution. 
In contrast to these extreme cases, there is,
e.g., $\RIO$ which incorporates both the diagnoses elimination rate and fault probabilities in its calculations. The consequence is a behavior that mostly lies in between the performances of $\SPL$ and $\ENT$. 
Based on the data in the figures, which is quite consistent in this regard, one could 
impose the following \emph{qualitative ordering from most to least probability-sensitive} on the QSMs: 
\begin{align}
\tuple{\EMCb, \BME, \ENT, \KL, \MPS, \RIO, \RND, \SPL} \label{eq:order_probability-sensitivity}
\end{align}

\noindent\textbf{Impact of the DPI / Diagnostic Structure.}
Trivially, the overall number of 
diagnoses to discriminate between impacts the average number of queries required. Thus, for M (48 diagnoses), U (90), E (864) and T (1782), respectively, the min/avg/max number of queries over all QSMs and sessions is (rounded) $3$/$7$/$18$, $4$/$8$/$19$, $6$/$10$/$19$ and $4$/$12$/$29$. The difference between M and E, for instance, can be quite well seen by comparing 
the length of the bars in 
Figures~\ref{fig:eval_minitambis} and \ref{fig:eval_economy} which are placed side by side. On the contrary and as one would expect, there are no indications of the system size $|\mo_j|$ (3rd column, Tab.~\ref{tab:dataset}) having a remarkable influence on QSM performance (as the system size has generally no bearing on the diagnoses number). The reasoning complexity (4th column, Tab.~\ref{tab:dataset}), in contrast, affects the query computation time, 
which was,
over all runs and QSMs, maximally $0.18$/$0.13$/$0.18$/$0.14$ sec (per query) for the cases M/U/E/T. The relative behavior of the QSMs under varying DPI (but otherwise same conditions) appears to be quite stable. To see this,
compare, e.g., the EQ, the MOD and the STR cases between Figures~\ref{fig:eval_university_overall} and \ref{fig:eval_transportation_overall}, or Figures~\ref{fig:eval_minitambis} and \ref{fig:eval_economy}. From the pragmatic point of view, if this consistency of QSM performances irrespective of the particular DPI generalizes (as needs to be verified using a larger dataset), a nice implication thereof would be the possibility to recommend (against) QSMs independently of (the structure of) the problem at hand. 

\noindent\textbf{Impact of the Leading Diagnoses.} As Figures~\ref{fig:eval_university_overall} and \ref{fig:eval_transportation_overall} indicate quite well, and numbers confirm, there is no significant average difference in the numbers of queries for varying $\ld \in \setof{6,10,14}$. This is in line with the findings of \cite{de1995trading}. What we can realize, though, is an exacerbation of the discrepancy between the plausible (green bars) and implausible (red bars) cases when $\ld$ increases. The random case (yellow bars), on the other hand, is mostly stable. The reason for this intensification of the effect of good or bad bias with larger diagnoses samples is that more extreme decisions might be made in this case. A simple illustration of this is to compare a ``risky'' \cite{Rodler2013} query (one that might invalidate very few diagnoses) wrt.\ a sample of $3$ and $100$ diagnoses; in the former case, this would be one eliminating either $1$ or $2$, 
in the latter 
one ruling out either $1$ or $99$ known hypotheses. We see that the former query is similar to a ``risk-less'' split-in-half choice, while the latter is far off being that conservative. A practical consequence of this is that it might make sense to try generating a higher number of diagnoses per iteration (if feasible in reasonable time) if a probability-based measure, e.g.\ $\EMCb$ or $\ENT$, is used and the trust in the given (biased) fault information is high (e.g.\ if reliable historical data is available). Verify this by considering $\EMCb$ and $\ENT$ in the MOD and STR cases for $\ld\in\setof{6,14}$ in Figures~\ref{fig:eval_university_overall} and \ref{fig:eval_transportation_overall}. By contrast, when adopting
a probability-insensitive QSM, say $\SPL$, one seems to be mostly better off when relying on a smaller $\ld$. That is, when the meta information is vague, a good option is to rely on a ``cautious'' \cite{Rodler2013} measure such as $\SPL$ \emph{and} a small diagnoses sample. Note, the latter is doubly beneficial as it also decreases computation times.

\renewcommand{\arraystretch}{1.2}
\begin{table*}[htbp]
	\setlength{\tabcolsep}{6.5pt}
	\scriptsize
	\centering
	\begin{tabular}{@{}cccccccccccc}
		\toprule
		& \multicolumn{3}{c}{PLAUSIBLE} & \phantom{.}      & \multicolumn{3}{c}{RANDOM} &  \phantom{.}     & \multicolumn{3}{c}{IMPLAUSIBLE} \\
		\cmidrule{2-4}\cmidrule{6-8}\cmidrule{10-12}          & EQ    & MOD   & STR   &       & EQ    & MOD   & STR   &       & EQ    & MOD   & STR \\
		\cmidrule{2-4}\cmidrule{6-8}\cmidrule{10-12}    \multirow{2}[1]{*}{M} & \underline{$\KL$} & \underline{$\RIO$}, $\ENT$, $\BME$ & \underline{$\MPS$}, $\ENT$ &       & \underline{$\MPS$} & \underline{$\MPS$} & \underline{$\MPS$} &       & \underline{$\MPS$} & \underline{$\RND$} & \underline{$\RND$} \\
		& \cellcolor[rgb]{ .529,  .784,  .49}63 & \cellcolor[rgb]{ .992,  .737,  .482}176 & \cellcolor[rgb]{ .996,  .839,  .502}144 &       & \cellcolor[rgb]{ .388,  .745,  .482}46 & \cellcolor[rgb]{ .396,  .745,  .482}47 & \cellcolor[rgb]{ .404,  .749,  .482}48 &       & \cellcolor[rgb]{ 1,  .922,  .518}118 & \cellcolor[rgb]{ 1,  .882,  .51}131 & \cellcolor[rgb]{ .973,  .412,  .42}277 \\
		\multirow{2}[0]{*}{U} & \underline{$\BME$} & \underline{$\BME$} & \underline{$\MPS$}, $\BME$ &       & \underline{$\MPS$} & \underline{$\MPS$} & \underline{$\MPS$} &       & \underline{$\RND$} & \underline{$\RND$} & \underline{$\RND$}, $\KL$ \\
		& \cellcolor[rgb]{ .804,  .863,  .506}59 & \cellcolor[rgb]{ .992,  .718,  .478}129 & \cellcolor[rgb]{ .988,  .643,  .467}151 &       & \cellcolor[rgb]{ .388,  .745,  .482}42 & \cellcolor[rgb]{ .58,  .8,  .49}50 & \cellcolor[rgb]{ .655,  .82,  .494}53 &       & \cellcolor[rgb]{ 1,  .922,  .518}67 & \cellcolor[rgb]{ .988,  .651,  .467}149 & \cellcolor[rgb]{ .973,  .412,  .42}220 \\
		\multirow{2}[0]{*}{E} & \underline{$\BME$} & \underline{$\ENT$} & \underline{$\EMCb$}, $\RIO$, $\BME$ &       & \underline{$\ENT$}, $\MPS$ & \underline{$\MPS$} & \underline{$\ENT$}, \underline{$\RIO$}, $\BME$, $\MPS$ &       & \underline{$\MPS$} & \underline{$\KL$} & \underline{$\RND$} \\
		& \cellcolor[rgb]{ .733,  .843,  .502}64 & \cellcolor[rgb]{ 1,  .91,  .518}93 & \cellcolor[rgb]{ 1,  .922,  .518}90 &       & \cellcolor[rgb]{ .388,  .745,  .482}30 & \cellcolor[rgb]{ .416,  .753,  .482}33 & \cellcolor[rgb]{ .459,  .765,  .486}37 &       & \cellcolor[rgb]{ .992,  .769,  .49}121 & \cellcolor[rgb]{ .973,  .412,  .42}191 & \cellcolor[rgb]{ 1,  .91,  .518}93 \\
		\multirow{2}[1]{*}{T} & \underline{$\EMCb$} & \underline{$\EMCb$}, $\RIO$, $\ENT$ & \underline{$\ENT$}, \underline{$\BME$}, \underline{$\EMCb$}, \underline{$\MPS$} &       & \underline{$\MPS$} & \underline{$\MPS$} & \underline{$\MPS$} &       & \underline{$\MPS$} & \underline{$\RND$} & \underline{$\RND$} \\
		& \cellcolor[rgb]{ .655,  .82,  .494}62 & \cellcolor[rgb]{ .992,  .722,  .482}125 & \cellcolor[rgb]{ .973,  .412,  .42}174 &       & \cellcolor[rgb]{ .463,  .765,  .486}45 & \cellcolor[rgb]{ .408,  .749,  .482}40 & \cellcolor[rgb]{ .388,  .745,  .482}38 &       & \cellcolor[rgb]{ 1,  .922,  .518}93 & \cellcolor[rgb]{ 1,  .867,  .51}102 & \cellcolor[rgb]{ .992,  .733,  .482}123 \\
		\bottomrule
	\end{tabular}%
	\caption{\small Shows which QSM(s) exhibited best performance in the various scenarios in (F1)$\times$(F3) for all DPIs (1st column) in Tab.~\ref{tab:dataset} and the setting $\ld = 10$ of (F4). The QSM(s) with lowest $\#$ of queries (per scenario) are underlined. All stated non-underlined QSMs lay within $3\%$ of the best QSM wrt.\ $\#$ of queries. The number below the QSM(s) gives the possible overhead $(\#q_{\mathit{worstQSM(S),S}} / \#q_{\mathit{bestQSM(S),S}}-1)*100$ in $\%$ incurred by using a non-optimal QSM in a scenario $S$, where $\#q_{X,S}$ refers to the $\#$ of required queries of QSM $X$ in scenario $S$, and $\mathit{bestQSM(S)}$ / $\mathit{worstQSM(S)}$ denote the best / worst QSM in scenario $S$. The colors signify criticality of QSM choice based on the overhead, from lowest=green to highest=red.}
	\label{tab:eval_best_QSM_per_setting}%
\end{table*}%

\noindent\textbf{Importance of Using a Suitable QSM.} To quantify the importance of QSM choice we compute the \emph{degree of criticality of choosing the right QSM in a scenario} as the overhead in oracle cost (number of queries) when employing the worst instead of the best QSM in this scenario, see (the caption of) Tab.~\ref{tab:eval_best_QSM_per_setting}. At this, a scenario is one factor level combination in (F1)$\times$(F3). 
We learn from Tab.~\ref{tab:eval_best_QSM_per_setting} that, even in the least critical cases (green-colored), we might experience a worst-case overhead in oracle effort of at least $30\%$ when opting for the wrong QSM. This overhead is drastically higher in other cases and reaches figures of over $250\%$. That is, more than triple the effort might be necessary to locate a fault under an inopportune choice of QSM heuristic. However, we emphasize that even a $30\%$ overhead must be considered serious given that usually oracle inquiries are very costly. Hence, 
appropriate QSM selection \emph{is} an important issue to be addressed in all scenarios.

As a predictor of the criticality, the scenario (columns in Tab.~\ref{tab:eval_best_QSM_per_setting}) appears to be a reasonable candidate, as the colors already suggest.
In fact, the coefficients of variance, one 
computed for each column in Tab.~\ref{tab:eval_best_QSM_per_setting}, are fairly low, ranging from $3\%$ to $26\%$ (except for the last column with $47\%$). So, the negative effect of a bad QSM choice is similar in equal scenarios, and does not seem to be dependent on the DPI.

\renewcommand{\arraystretch}{1}
\begin{table}[t]
	\centering
	\setlength{\tabcolsep}{2.7pt}
	\scriptsize
	\begin{tabular}{@{}llrrrrrrrr@{}}
		& 
		& \multicolumn{1}{l}{$\ENT$} 
		& \multicolumn{1}{l}{$\SPL$} 
		& \multicolumn{1}{l}{$\KL$} 
		& \multicolumn{1}{l}{$\EMCb$} 
		& \multicolumn{1}{l}{$\MPS$} 
		& \multicolumn{1}{l}{$\BME$} 
		& \multicolumn{1}{l}{$\RIO$} 
		& \multicolumn{1}{l}{$\RND$} \\
		\toprule
		\multirow{2}[2]{*}{ALL} & among best & 7     & 0     & 3     & 4     & 18    & 8     & 4     & 8 \\
		 & \emph{the} best & 4     & 0     & 2     & 4     & 16    & 4     & 2     & 8 \\
		 \midrule
		 \multirow{2}[2]{*}{PLAUSIBLE} & among best & 5     & 0     & 1     & 4     & 3     & 7     & 3     & 0 \\
		 & \emph{the} best & 2     & 0     & 1     & 4     & 3     & 4     & 1     & 0 \\
		\bottomrule
	\end{tabular}%
	\caption{\small Number of times each QSM is (among) the best in Tab.~\ref{tab:eval_best_QSM_per_setting}.}
	\label{tab:eval_how_often_is_QSM_best}%
\end{table}%

\noindent\textbf{Which QSM to use in which Scenario?} To approach this question, we have, for all four DPIs, analyzed all the nine settings in (F1)$\times$(F3) regarding the optimal choice of a QSM. The result is presented in Tab.~\ref{tab:eval_best_QSM_per_setting}. We now discuss various insights from this analysis.

\noindent\emph{Overall Picture.} $\SPL$ is never a (nearly) optimal option. This is quite natural because, intuitively, going for no ``risk'' at all means at the same time excluding the chance to perform extraordinarily well. All other QSMs appear multiple times among those QSMs which are $\leq 3 \%$ off the observed optimal number of queries. Tab.~\ref{tab:eval_how_often_is_QSM_best} (rows 1+2) lists how often each QSM is (among) the best. It shows that $\MPS$ is close to the optimum in a half of the cases, significantly more often than all other heuristics. However, blindly
deciding for $\MPS$ is not a rational way to go. Instead, one must consider the numbers at a more fine-grained level, distinguishing between the quality 
of the given fault distribution (blocks in Tab.~\ref{tab:eval_best_QSM_per_setting}), to get a clearer and more informative picture.

\noindent\emph{The Implausible Cases:} Here $\RND$ distinctly prevails. It occurs in all but four optimal QSM sets, and is often \emph{much} better than other measures, e.g., see the STR setting in Fig.~\ref{fig:eval_minitambis}. At first sight, it might appear counterintuitive that a random selection outweighs all others. One explanation is simply that the randomness prevents $\RND$ from getting misled by the (wrong) fault information. Remarkable is, however, that in quasi all cases $\RND$ significantly outperforms $\SPL$, which 
acts independently of the given probabilities as well.
The conclusion from this is that, whenever the prior distribution is wrongly biased, introducing randomness into the query selection procedure saves oracle effort.

\noindent\emph{The Random Cases:} These cases are strongly dominated by $\MPS$ which occurs in each set of best QSMs per scenario. Therefore, whenever the given fault information does neither manifest a tendency towards nor against the actual diagnosis, $\MPS$ is the proper heuristic. Moreover, the benefit of using $\MPS$ seems to increase the more leading diagnoses are available for query selection (see Figures \ref{fig:eval_university_overall} and \ref{fig:eval_transportation_overall}). Since $\MPS$, in attempt to invalidate \emph{a maximal number of diagnoses}, suggests very ``risky'' queries (see above), a possible explanation for this is that acting on a larger diagnoses sample allows to guarantee a higher risk than when relying on a smaller sample (cf.\ discussion above). 
However, as all Figures \ref{fig:eval_university_overall} -- \ref{fig:eval_economy} clearly reveal, $\MPS$ is definitely the wrong choice in any situation where we have a plausible, but unbiased probability distribution. In such cases it manifests sometimes significantly worse results than other heuristics do. But, as soon as a bias is given, the performance of MPS gets really good.

\noindent\emph{The Plausible Cases:} Throughout these cases we have the highest variation concerning the optimal QSM. Actually, all QSMs except for $\RND$ and $\SPL$ do appear as winners in certain cases. 
The distribution of the number of appearances as (or among) the best QSM(s) over all QSMs is displayed by Tab.~\ref{tab:eval_how_often_is_QSM_best} (rows 3+4). That, e.g., $\ENT$ is rather good in these cases and $\RND$ is no good choice (see also the Figures \ref{fig:eval_university_overall} -- \ref{fig:eval_economy}) is in agreement with the findings of \cite{Shchekotykhin2012}. However, we realize that $\BME$ is (among) the best QSMs more often than $\ENT$. Comparing only these two, we find that $\BME$ outdoes $\ENT$ $7$ times, $\ENT$ wins against $\BME$ $4$ times, and they are equally good once. 
A reason for the strength of $\BME$ could be the fact that it will in most cases achieve only a minor bias towards one query outcome, as the maximization of the diagnoses elimination rate requires an as small as possible number of diagnoses with a probability sum $>0.5$. Thence, there is on the one hand a bias increasing the expected diagnoses invalidation rate, and on the other hand a near 50-50 outcome distribution implying a good entropy value of the query. 
Unsurprisingly, if we sort the QSMs from most to least times being (among) the best based on Tab.~\ref{tab:eval_how_often_is_QSM_best} (rows 3+4), the resulting order coincides quite well with Eq.~\eqref{eq:order_probability-sensitivity}. In other words, in the plausible scenarios, probability-sensitive heuristics perform best.

\noindent\textbf{Towards new QSMs / Meta-Heuristics.}
Exploiting the discussed results, one could endeavor to devise new QSMs that are superior to the investigated ones.
For instance, in the implausible cases, 
only $\RND$, $\MPS$ and $\KL$ occur as best QSMs. Thus, an optimal heuristic for these cases should likely adopt or unify selection principles of these three QSMs. One idea could be, e.g., to sample a few queries using $\RND$ and then choose the best one among them using (a weighted combination of) $\MPS$ and/or $\KL$. Generally, one could use a meta heuristic that resorts to an appropriately (possibly dynamically re-)weighted sum of the QSM-functions (Tab.~\ref{tab:QSMs}, 2nd column).
Also, a QSM selecting queries based on a majority voting of multiple heuristics is thinkable, e.g., in Ex.~\ref{ex:QSMs} the query selected by such a QSM would be $q_7$.

\section{Conclusions}
\label{sec:conclusion}
Results of extensive evaluations on both classical and recently suggested query 
selection measures (QSMs) for sequential diagnosis (SD) are presented. Main findings are: Using an appropriate QSM is essential, as otherwise SD cost overheads of over $250\%$ are possible. The one and only best QSM does not exist (or has not yet been found). Besides the size of the solution space, main factors influencing SD cost are the bias in and the quality of the fault probability distribution, but not the diagnosis problem as such or the size of the diagnoses sample used for query selection. Different QSMs prevail in the various probability distribution scenarios. Interestingly, the very popular entropy measure only manifested good (albeit not best) behavior in a single set of scenarios.

Future work topics include in-depth analyses of the (full) results and the design of new QSMs, e.g.\ meta-heuristics, based on the lessons learned. 
Moreover, machine learning techniques could be adopted to recommend optimal QSMs based on a classification of a diagnosis scenario wrt.\ the QSM-relevant factors we found. 
%
And, we plan to integrate the investigated QSMs into our \protege ontology debugging plug-in.\footnote{http://isbi.aau.at/ontodebug/plugin}  



\begin{landscape}
	\flushleft
	\begin{minipage}{0.48\linewidth}
		
		\centering
		\includegraphics[width=12.1cm]{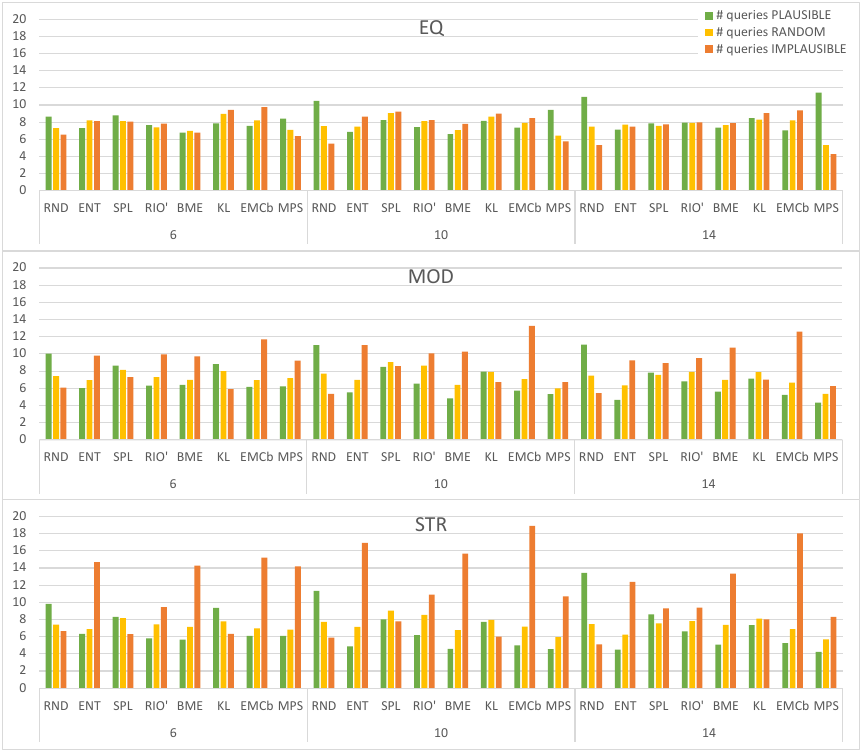}
		\captionof{figure}{\small Experimental results for the University (U) case.}
		\label{fig:eval_university_overall}
		
	\vspace{20pt}		
		
		\centering
		\includegraphics[width=12.1cm]{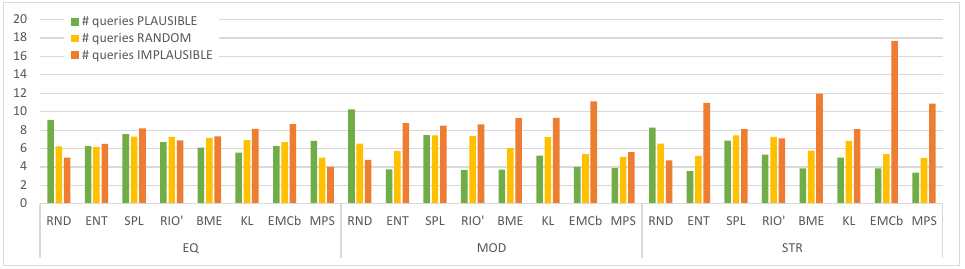}
		\captionof{figure}{\small Experimental results for the MiniTambis (M) case with $\ld = 10$ (F4).}
		\label{fig:eval_minitambis}
		
	\end{minipage}
	\hfill
	\begin{minipage}{0.48\linewidth}
		
		\centering
		\includegraphics[width=12.5cm]{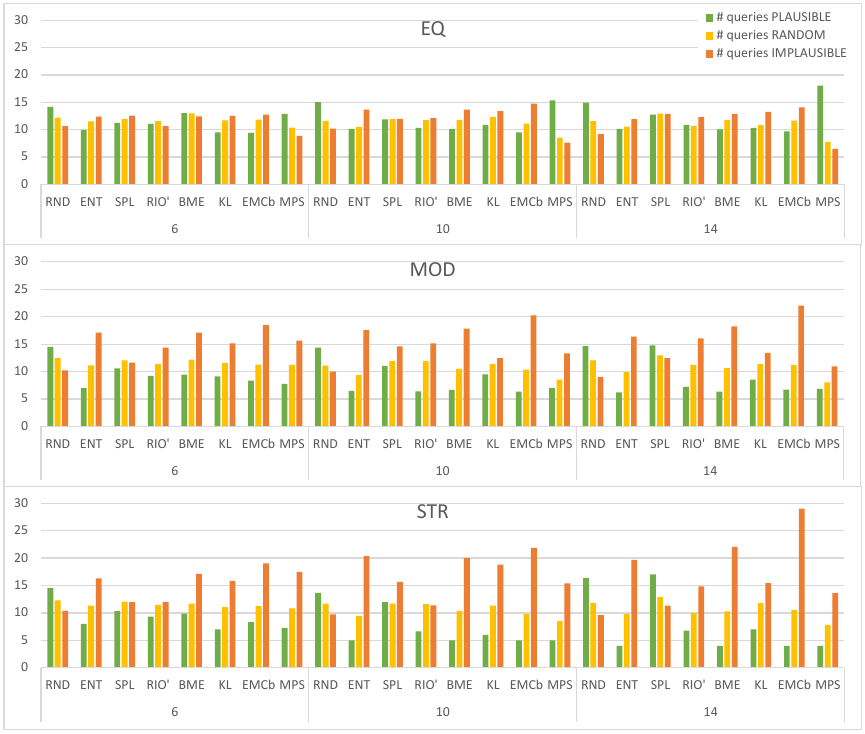}
		\captionof{figure}{\small Experimental results for the Transportation (T) case.}
		\label{fig:eval_transportation_overall}
		
	\vspace{20pt}	
		
		\centering
		\includegraphics[width=12.5cm]{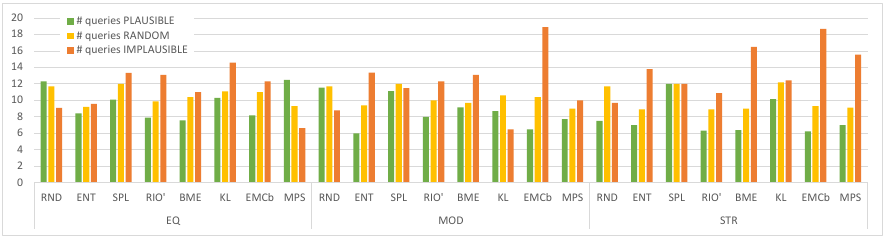}
		\captionof{figure}{\small Experimental results for the Economy (E) case with $\ld = 10$ (F4).}
		\label{fig:eval_economy}
		
	\end{minipage}
\end{landscape}

\end{document}